\pdfoutput=1

\documentclass[11pt]{article}

\usepackage[final]{ACL2023}

\usepackage{times}
\usepackage{latexsym}

\usepackage[T1]{fontenc}

\usepackage[utf8]{inputenc}

\usepackage{microtype}

\usepackage{inconsolata}

\usepackage{booktabs}
\usepackage{todonotes}
\usepackage{graphicx}
\usepackage{xspace}
\usepackage{hyperref}
\usepackage{subcaption}

\definecolor{googlemakeryellow}{HTML}{FEE69A}
\newcommand{\yellowmark}[1]{{\setlength{\fboxsep}{2pt}\hspace{-2pt}\colorbox{googlemakeryellow}{#1}}}

\title{Preconditioned Visual Language Inference with Weak Supervision}

\usepackage{cleveref}
\crefformat{section}{\S#2#1#3}
\crefformat{subsection}{\S#2#1#3}
\crefformat{subsubsection}{\S#2#1#3}
\crefrangeformat{section}{\S\S#3#1#4 to~#5#2#6}
\crefmultiformat{section}{\S\S#2#1#3}{ and~#2#1#3}{, #2#1#3}{ and~#2#1#3}
\usepackage{refstyle}
\usepackage{graphicx}
\Crefformat{figure}{#2Fig.~#1#3}
\Crefmultiformat{figure}{Figs.~#2#1#3}{ and~#2#1#3}{, #2#1#3}{ and~#2#1#3}
\Crefformat{table}{#2Tab.~#1#3}
\Crefmultiformat{table}{Tabs.~#2#1#3}{ and~#2#1#3}{, #2#1#3}{ and~#2#1#3}
\Crefformat{appendix}{Appx.~\S#2#1#3}
\crefformat{algorithm}{Alg.~#2#1#3}

\newcommand{\CSn}{common sense\xspace}

\newcommand{\QT}[1]{``{#1}''\xspace}

\definecolor{USCgold}{HTML}{F6C400}
\newcommand{\pvli}{\textit{PVLIR}\xspace}

\newcommand\blfootnote[1]{%
  \begingroup
  \renewcommand\thefootnote{}\footnote{#1}%
  \addtocounter{footnote}{-1}%
  \endgroup
}

\author{Ehsan Qasemi{$^{\dagger}$} 
    \and Amani R. Maina-Kilaas {$^{1}$} 
    \and Devadutta Dash {$^{2}$} \\
    \and \textbf{Khalid Alsaggaf} {$^{3}$} 
    \and \textbf{Muhao Chen}{$^{\dagger}$} 
    \\
    \textbf{$\dagger$}Department of Computer Science, University of Southern California
    \\ 
    \textbf{$^1$}Harvey Mudd College, 
    \textbf{$^2$}Indian Institute of Technology, 
    \textbf{$^3$}University of Wisconsin-Madison \\ 
    \{qasemi,muhaoche\}@usc.edu, amainakilaas@g.hmc.edu, devadutta.dash.cd.cse19@itbhu.ac.in\\
    alsaggaf@wisc.edu\\}

\begin{document}

\maketitle
\blfootnote{$1$ Work performed during REU Program at USC-ISI.}
\blfootnote{$2$ Work performed during IUSSTF Viterbi India Program Internship at USC-ISI.}
\blfootnote{$3$ Work performed during IKAUST Exchange Student Internship at USC-ISI.}
\begin{abstract}
    Humans can infer the affordance of objects by extracting related contextual preconditions for each scenario. 
    For example, upon seeing an image with a broken cup, we can infer that this precondition prevents the cup from being used for drinking. 
    Reasoning with preconditions of commonsense is studied in NLP where the model explicitly gets the contextual precondition. 
    However, it is unclear if SOTA visual language models (VLMs) can extract such preconditions and infer the affordance of objects with them. 
    In this work, we introduce the task of preconditioned visual language inference and rationalization (PVLIR). 
    We propose a learning resource based on three strategies to retrieve weak supervision signals for the task and develop a human-verified test set for evaluation.
    Our results reveal the shortcomings of SOTA VLM models in the task and draw a road map to address the challenges ahead in improving them.
\end{abstract}
\section{Introduction}
\label{sec:introduction}
According to the \emph{Theory of Affordance}~\cite{affordance,chemero2003outline}, understanding the preconditions in which an action or statement is possible or impossible is a key aspect of human intelligence.
For example, a glass may be used for drinking water, under an implicit assumption that the water is at normal temperature, but may not be if the glass is shattered.
From the cognitive perspective, understanding the affordance of objects, or simply preconditions of actions~\cite{qasemi2022paco}, is part of the commonsense knowledge that constitutes what distinguishes humans from a machine to make inference~\cite{lenat1998dimensions}.
From an applications perspective, it also has huge implications such as robotics~\cite{ahn2022can}, transportations~\cite{prakken2017problem,seff2016learning,kothawade2021auto}, and general artificial intelligence~\cite{nguyen2021advanced}.
\begin{figure}[t]
    \centering
    \includegraphics[width=\linewidth]{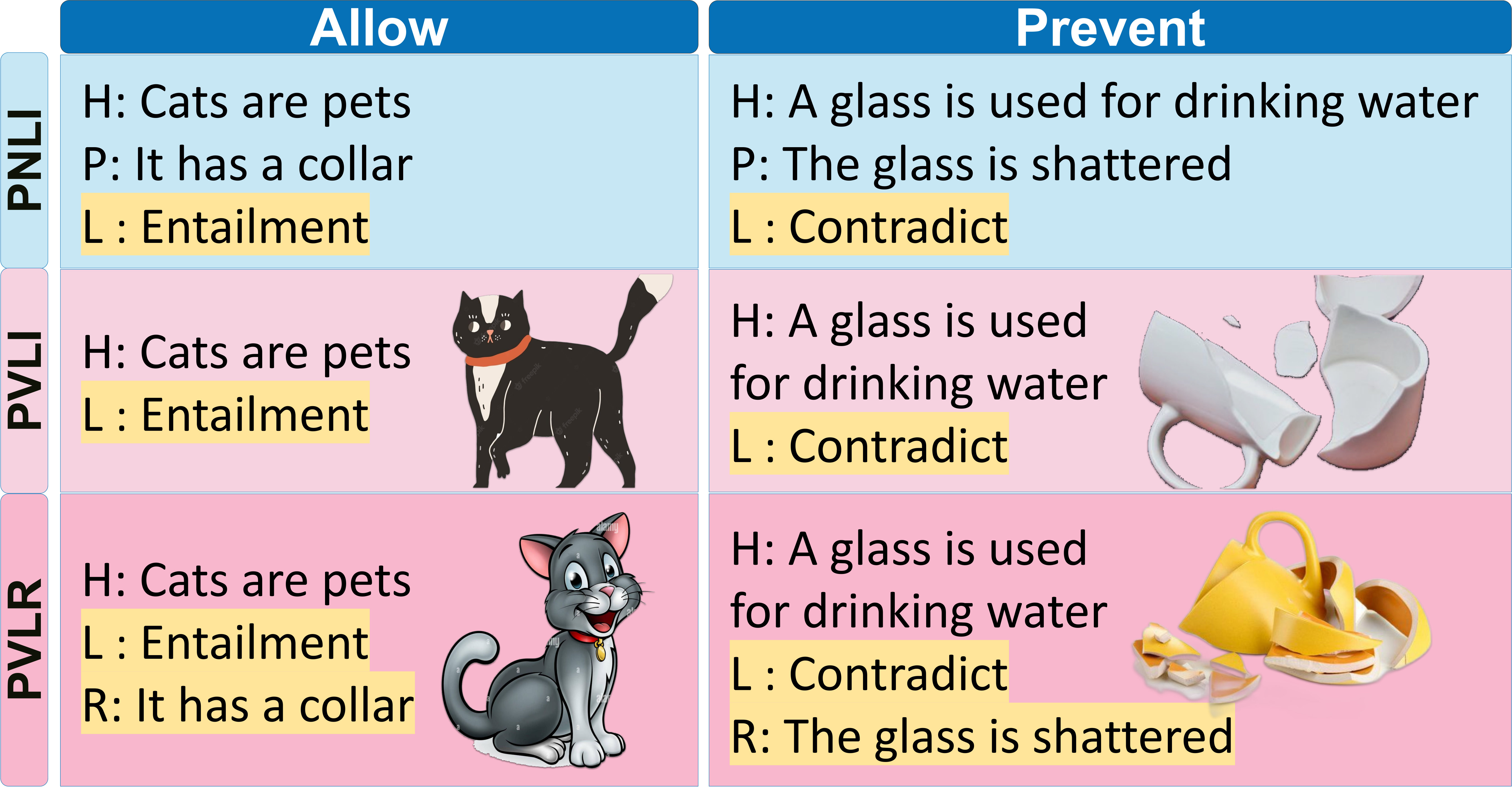}
    \caption{Preconditioned Visual Language Inference (PVLI) and Preconditioned Visual Language Reasoning (PVLR) tasks. The \QT{H} and \QT{P} are the input \textit{hypothesis} and \textit{premise}. The outputs, \textit{label} (letter \QT{L}) and \textit{rationale} (letter \QT{R}), are \yellowmark{highlighted}.}
    \label{fig:pvlr-pvli}
    \vspace{-1.5em}
\end{figure}

Reasoning with preconditions of commonsense knowledge (i.e. preconditioned inference), is proposed as a benchmarking task for evaluation of the theory of affordance~\cite{qasemi2022paco}. 
Multiple studies have formulated the preconditioned natural language inference (PNLI) as variations of the Natural Language Inference (NLI) ~\cite{mnli,snli,condoravdi2003entailment} task and contributed learning resources that are gathered through crowdsouring~\cite{rudinger2020thinking,qasemi2022paco,hwang2020comet,do2021rotten,jiang2021mad} or weak supervision data~\cite{qasemi2022pinks}.
In PNLI, the models rely on the  contextual information (i.e. textual preconditions as \textit{premise}) as input and have to decide whether the \textit{hypothesis} is true (entailment), false (contradiction), or undetermined (neutral) given the \textit{premise} (first row in \Cref{fig:pvlr-pvli}).
However, humans reason about affordance using information beyond text~\cite{barsalou2010grounded,andrews2009integrating} and extract the contextual meaning representations for cognitive tasks (such as PNLI) from the pool of available information in various modalities.   
For example, upon getting the query \QT{can this person run?} and seeing a picture of a person in a full leg cast, one can imply the contextual information from the image that \QT{the person is injured and incapable of running} and use it to answer the query accordingly.
Thus, a visual variation of the PNLI task is cognitively more realistic to benchmark artificial intelligence models.

In this work, we propose \pvli, to expand the preconditioned inference and reasoning to the visual-language realm by considering the interaction between linguistic and visual information in common sense.
This work presents three contributions.
\textbf{First}, we introduce the Preconditioned Visual Language Inference (PVLI) and Rationalization (PVLR) tasks (2nd and 3rd rows in \Cref{fig:pvlr-pvli}), which evaluate the visual-language models' (VLM) capabilities to reason with preconditions associated with commonsense knowledge.
In PVLI, the precondition is represented as an image
that further constrains the context in which the model has to decide the \QT{\textit{prevent}} or \QT{\textit{allow}} labels.
In PVLR, the model has to provide the rationale for the choice between the labels as well.
For example, say the model is given a commonsense statement such as \QT{a glass is used for drinking water} and/or an image of \QT{a person drinking water} as the \textit{hypothesis} and an image of a \QT{broken glass} as the \textit{premise}. Then, in PVLI, the model has to decide whether there is a \textit{prevented} or \textit{allowed} relation between them, and in PVLR, it has to provide a rationale for its decision, such as \textit{the glass is broken}.
In addition, to foster further research, we created a crowd-verified evaluation dataset to benchmark future models.

\textbf{Second}, we propose three strategies for retrieving a rich amount of cheap and allowably noisy supervision signals for inference and rationalization.
Similar to \citet{parcalabescu2021valse}, \pvli's three strategies rely on the available image captioning datasets (e.g. \citet{changpinyo2021cc12m,sharma2018conceptual,gurari2020captioning,lin2014microsoft}) that are readily available as a result of years of research in the field and maturity of resources.
In the first strategy, \emph{Extraction from Captions}, we utilize the PInKS~\cite{qasemi2022pinks} method to extract PNLI instances from image captions.
PInKS uses a combination of linguistic patterns (e.g. \QT{\{action\} unless \{precondition\}}) and generative augmentation to extract large quantities of instances from raw text.
In the second strategy, \emph{Caption Querying}, we use the existing crowdsourced PNLI instances (e.g. \citet{rudinger2020thinking,qasemi2022paco,hwang2020comet,do2021rotten,jiang2021mad}) and find an image caption that is semantically identical to them.
The third strategy, \emph{Image Querying}, focuses solely on the PNLI instances and devises queries (such as \QT{you are in a desert}) to search directly for corresponding images on the web using image search engines (e.g. \hyperlink{https://images.google.com/}{Google Images}).

Our \textbf{third} contribution is an extensive benchmarking of VLMs based on \pvli.
We benchmark 4 SOTA VLMs, FLAVA~\cite{singh2022flava}, VisualBERT~\cite{li2019visualbert}, ViLBERT~\cite{lu2019vilbert} and ViLT~\cite{kim2021vilt} in inference (\Cref{subsec:pvli-results}) and \citet{ayyubi2020generating} in rationalization (\Cref{subsec:pvlr-for-pvli}).
We show how an effective rationalization will improve inference in the VLM models~(\Cref{subsec:pvlr-for-pvli}).
We further investigate the fine-tuning (learning) process of VLMs in the inference task~(\Cref{subsec:curve-results}) and study their exploitation of the spurious correlation in our dataset~(\Cref{subsec:bias-results}).


\section{Construction of \pvli and Test Set}
\label{sec:dataset}
\label{subsec:dataset-methodology}

\begin{figure*}[t]
    \centering
    \includegraphics[width=\linewidth]{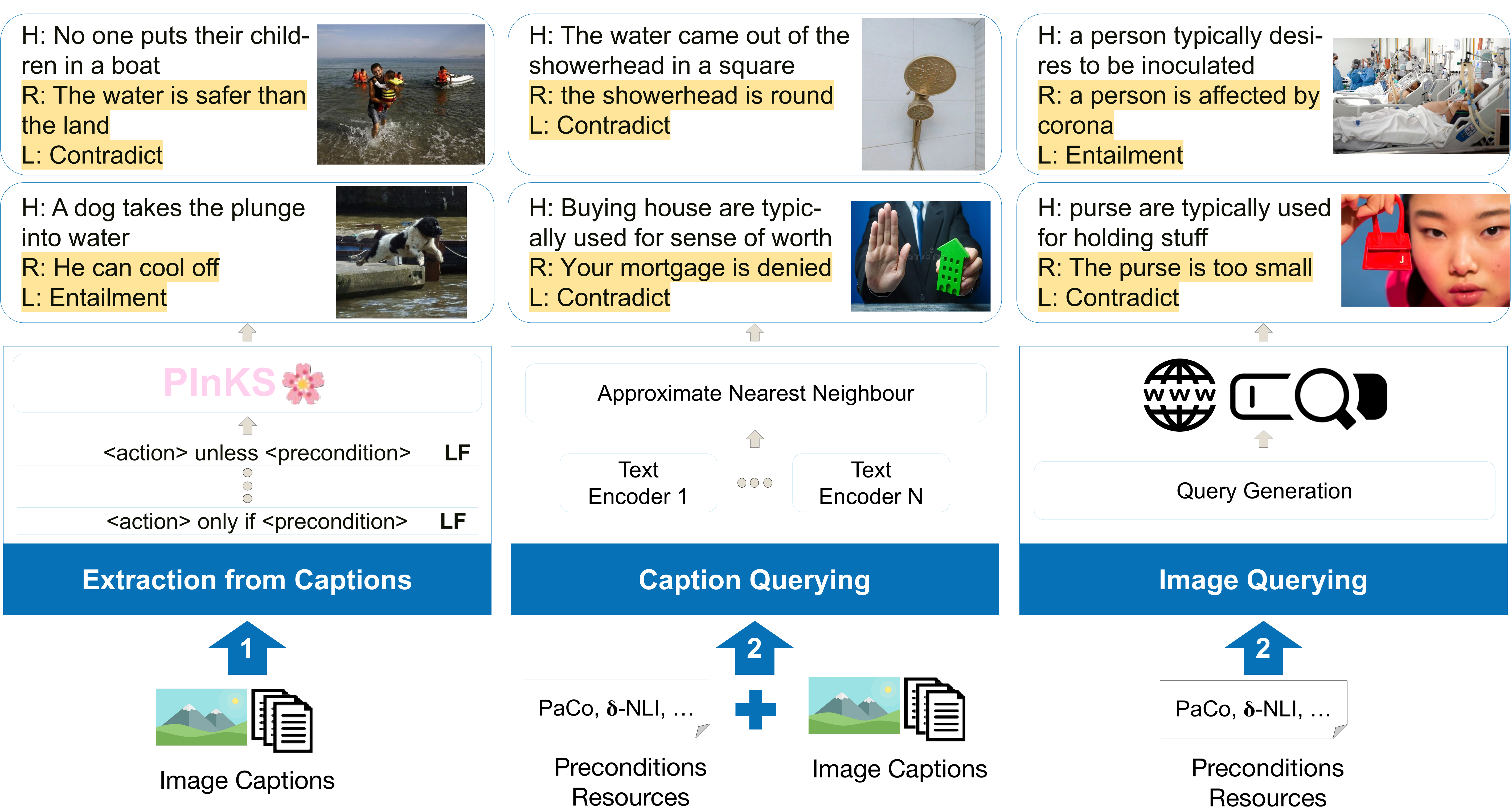}
    \caption{Overview of weak supervision methods for constructing \pvli.}
    \label{fig:weak-overview}
    \vspace{-1.5em}
\end{figure*}

This section gives an overview of \pvli (summarized in \Cref{fig:weak-overview}), describing our strategies for obtaining the data, and quality control.
Details associated with generating the human-verified test subset are moved to \Cref{sec:appendix-amt}.
The implementation details of each strategy are discussed in \Cref{subsec:data-experimental-setup}.

\paragraph{Datasets:}
The construction of \pvli uses existing text-only PNLI and image-captioning datasets as building blocks. 
For the text-only PNLI datasets, we require that they contain a precondition (e.g. premise, context), an action (e.g. hypothesis, question), and a binary label indicating whether the precondition allows or prevents the action. For the image captioning datasets, we simply require images (typically the URL) and their captions. Any datasets that meet these requirements can be used for the following steps.

\paragraph{Preprocessing:}

The PNLI preconditions and actions (collectively referred to as statements) often use varying conventions for referring to people. 
We standardize them by replacing these identifiers with \QT{the person}, \QT{another person}, \QT{a third person}, and so on.
For example, the sentence \QT{Alice helps Bob} would become \QT{the person helps another person}. This ensures that the specific names or traditionally-associated genders are not mistaken as a focus of the statements.
We also encourage further preprocessing of the data as seen fit, as not all sources contain clean text.
Of the image captions, some may be very long and consist of multiple sentences. 
In these cases, we split the captions into individual examples, pairing each sentence with the same image. 
Using these preprocessed resources, we then obtain \pvli instances with three different strategies: extraction from captions, caption querying, and image querying.

\paragraph{Extraction from Captions (EC):}

Our first strategy focuses solely on the image captions, finding the few that contain preconditions and actions and extracting them. 
By nature, the resulting statements are already grounded in the images.
We use the minimally-supervised approach described in \textit{PInKS}~\cite{qasemi2022pinks}, where linguistic patterns are used to extract preconditions and actions from raw corpora. 
This strategy constructs \textit{labeling functions (LF)} based on common conjunctions such as \QT{only if} and \QT{unless}. 
For example, the sentence \QT{Swimming pools have cold water in the winter unless they are heated} is matched by the pattern \QT{\{action\} unless \{precondition\}}, and therefore we can infer that \QT{they are heated} is a precondition that prevents the action \QT{Swimming pools have cold water in the winter}. 
Such labeling functions can be refined and added to as desired. 
In cases where the conjunction can be used in multiple senses, part-of-speech tagging can be utilized to filter out irrelevant senses.
After applying the labeling functions to the image captions, we have a dataset consisting of preconditions and actions, where both are grounded in the associated images.
To control for quality, we annotate a sample of matches from each labeling function (precision of each LF) to record whether the relation between precondition and action makes sense. 
Based on the results, we choose a precision threshold and only include labeling functions that meet this minimum.

\paragraph{Caption Querying (CQ):}

Our second strategy bridges the PNLI statements and image captions by grounding preconditions and actions in images that have semantically similar captions. 
We begin by limiting the statements and captions to those whose length is within one standard deviation of the mean (rounded to the nearest integer) in order to remove outliers. 
We then encode the PNLI statements and image captions in high-dimensional vector embeddings using multiple models.
Next, using a PNLI precondition as a query, we find the most similar captions through approximate nearest neighbors. 
This returns multiple rankings of the closest captions, one for each model's encoding. 
We then aggregate the rankings and select the first-place caption. 
This strategy of including multiple models in the decision-making process helps make it more robust to model differences and to the approximate nature of the nearest neighbors. 
The number of models incorporated depends on balancing the desired robustness and time or computational constraints. 
Likewise, the number of similar captions returned for the rankings can vary but should be chosen such that the rankings typically contain some overlap.
Note that the quality of data produced from this approach is dependent on the range of concepts covered by the image caption datasets, as it assumes that each PNLI precondition has a fairly similar caption. 
As such, it benefits from a very large corpus of captions (more discussion in \Cref{sec:limitations}).

To control for quality, when we select the best caption for a given query, we additionally record two values: \textit{perplexity} and \textit{model agreement}.
The perplexity is the distance (cosine, dot, etc.) between the query and caption, averaged over the models.
In the case when one of the models did not include the chosen caption in its ranking, the distance of the last caption is used for the average. 
By nature, the perplexity measures how good the models believe the match to be.
In contrast, the model agreement is not specific to the chosen caption but instead measures how well-aligned the
models' rankings are. 
Using a ranking similarity metric, we compute the similarity between pairs of rankings and then average the scores for the model agreement. 
Since a high model agreement indicates that the models agree on which are the closest captions, but does not speak to the actual proximity of the match, it can be thought of as a measure of confidence.

\paragraph{Image Querying (IQ):}

Our third strategy focuses solely on the PNLI statements and utilizes advances in image search engines to directly find the relevant images on the internet. 
Like the caption querying strategy, we limit the statements to those whose length is within one standard deviation of the mean.
For this approach, we recommend excluding any PNLI datasets that deal with largely abstract concepts (e.g. the person is responsible, the person will be grateful), as searching directly for images is unlikely to yield good results.
To ground the PNLI statements, we find the top images on the internet, using the statement with any commas removed as the search query. 
Since each of the top images becomes its own example, this strategy can quickly generate very large amounts of noisy training data. 
\section{Data Analysis}
\label{sec:data-analysis}
\label{sec:weak-results}

In this section, we investigate different aspects of the weak supervision data and evaluate the quality of the generated resource.
Implementation details and experimental setup details are moved to \Cref{subsec:data-experimental-setup} (for data acquisition), and \Cref{sec:appendix-amt} (for human annotations) to conserve space.

\begin{figure}[ht]
    \centering
    \includegraphics[width=\linewidth]{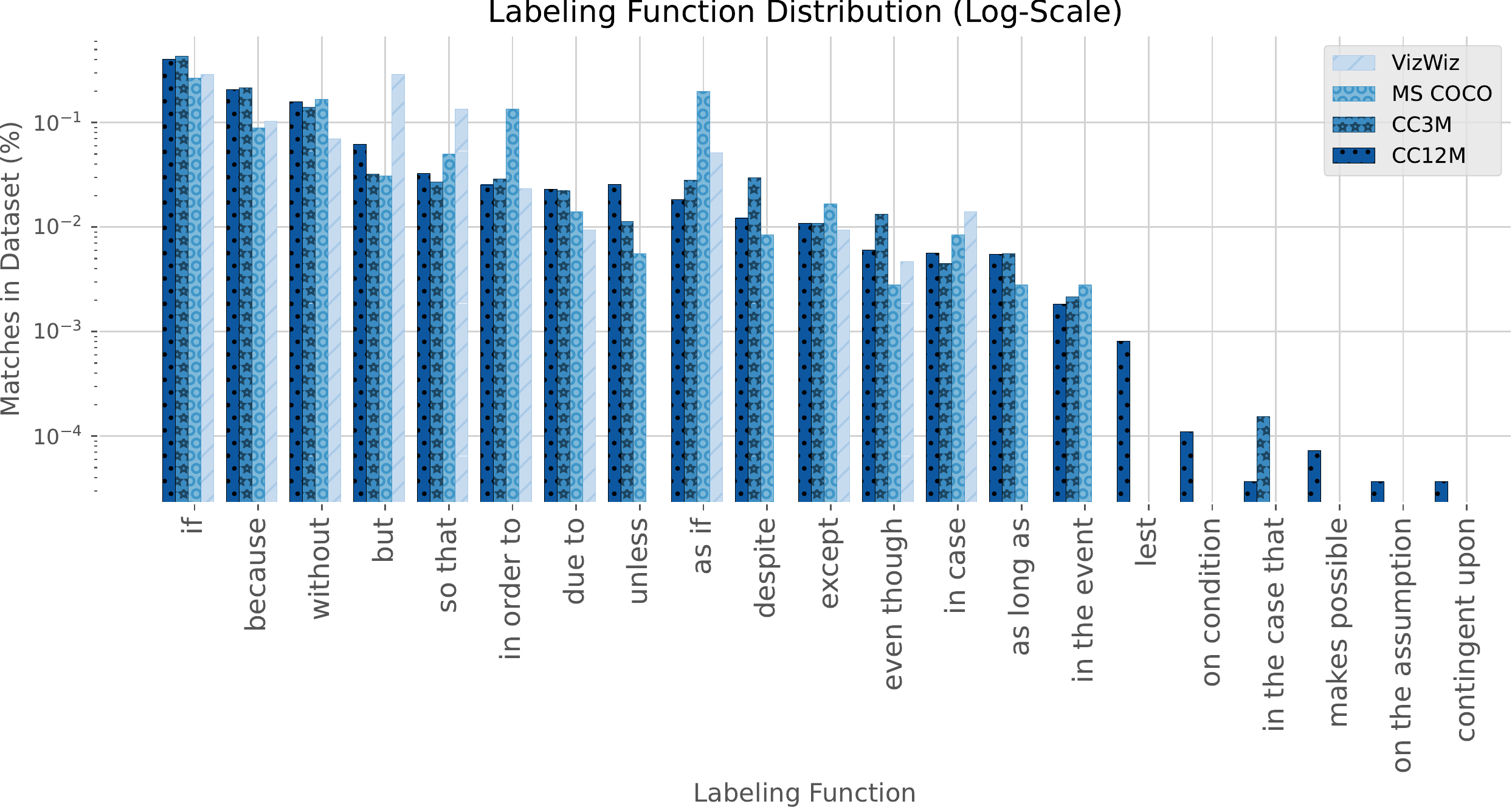}
    \caption{Distribution of instances extracted from captions (log-scale), for each source of the caption.}
    \label{fig:labeling-function-distribution}
\end{figure}

\begin{figure}[ht]
    \centering
    \includegraphics[width=0.9\linewidth]{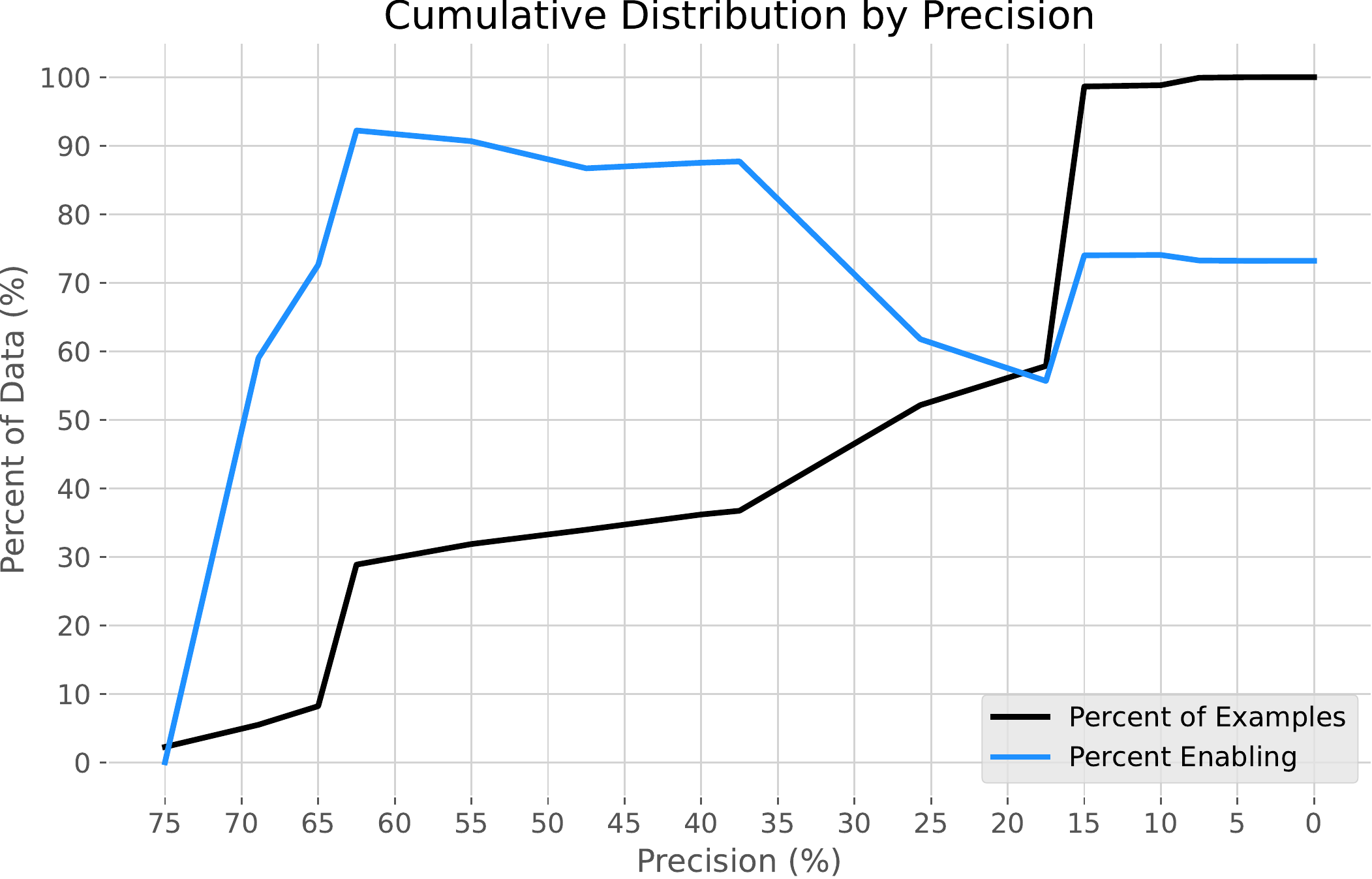}
    \caption{Cumulative distribution of the data with respect to the precision of the generating labeling function. 
    }
    \label{fig:cumdist}
\end{figure}

We then use a random sample of 3\% of the preprocessed preconditions and actions to perform \textit{Caption Querying} (roughly 13k each) and \textit{Image Querying} (roughly 80k each) to provide a representative sample of the full dataset. 

\paragraph{Extraction from Captions Results:} 
After preprocessing, we have a resource of 17 million captions.
From this, we utilize the \textit{Extraction from Captions} method that results in 34K extracted instances. 
\Cref{fig:labeling-function-distribution} illustrates the percentage of matches that come from each labeling function, separated by dataset. General statements such as \QT{if} unsurprisingly make up a large percentage of the data, but interestingly, some datasets have very different distributions. 
Among the sources of captions, VizWiz has disproportionately high counts of \QT{but} and \QT{so that}, while MS COCO is high in \QT{in order to} and \QT{as if}.

\Cref{fig:cumdist} shows the percent of the data and the percent of \QT{allow} examples for varying precision thresholds. 
For the results in \Cref{sec:vlm-results}, we use the threshold of $0.6$ to have a good balance between the quality and quantity of the final resource.

\begin{figure*}[ht]
\centering
\hspace*{\fill}
\begin{subfigure}[b]{0.45\textwidth}
    \centering
    \includegraphics[width=0.7\columnwidth]{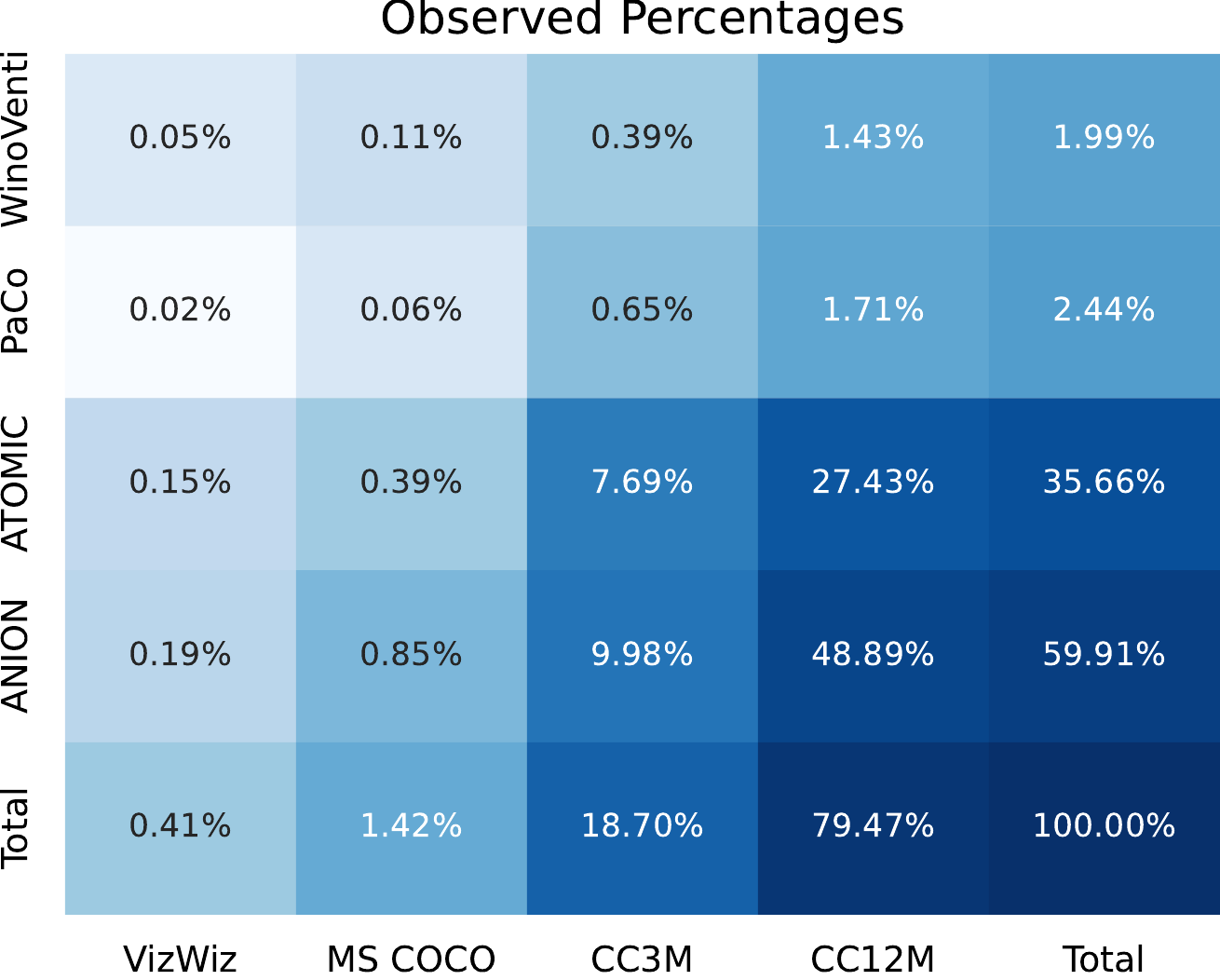}
    \caption{}
    \label{fig:preconditions_percents}
 \end{subfigure}
 \hfill
 \begin{subfigure}[b]{0.45\textwidth}
    \centering
    \includegraphics[width=.7\linewidth]{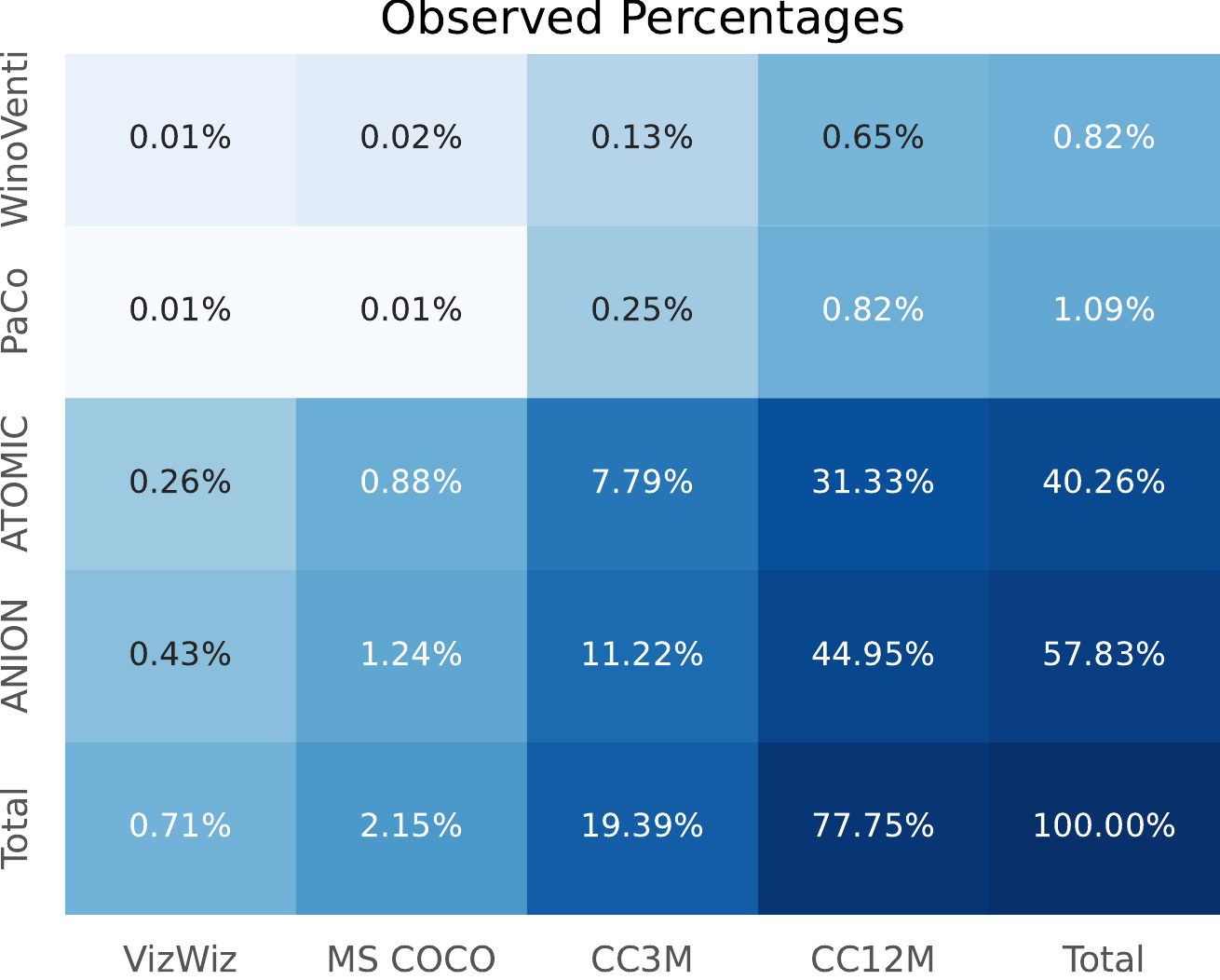}
    \caption{}
    \label{fig:actions_percents}
 \end{subfigure}
 \hspace*{\fill}
 \\
 \hspace*{\fill}
 \begin{subfigure}[b]{0.45\textwidth}
    \centering
    \includegraphics[width=.7\linewidth]{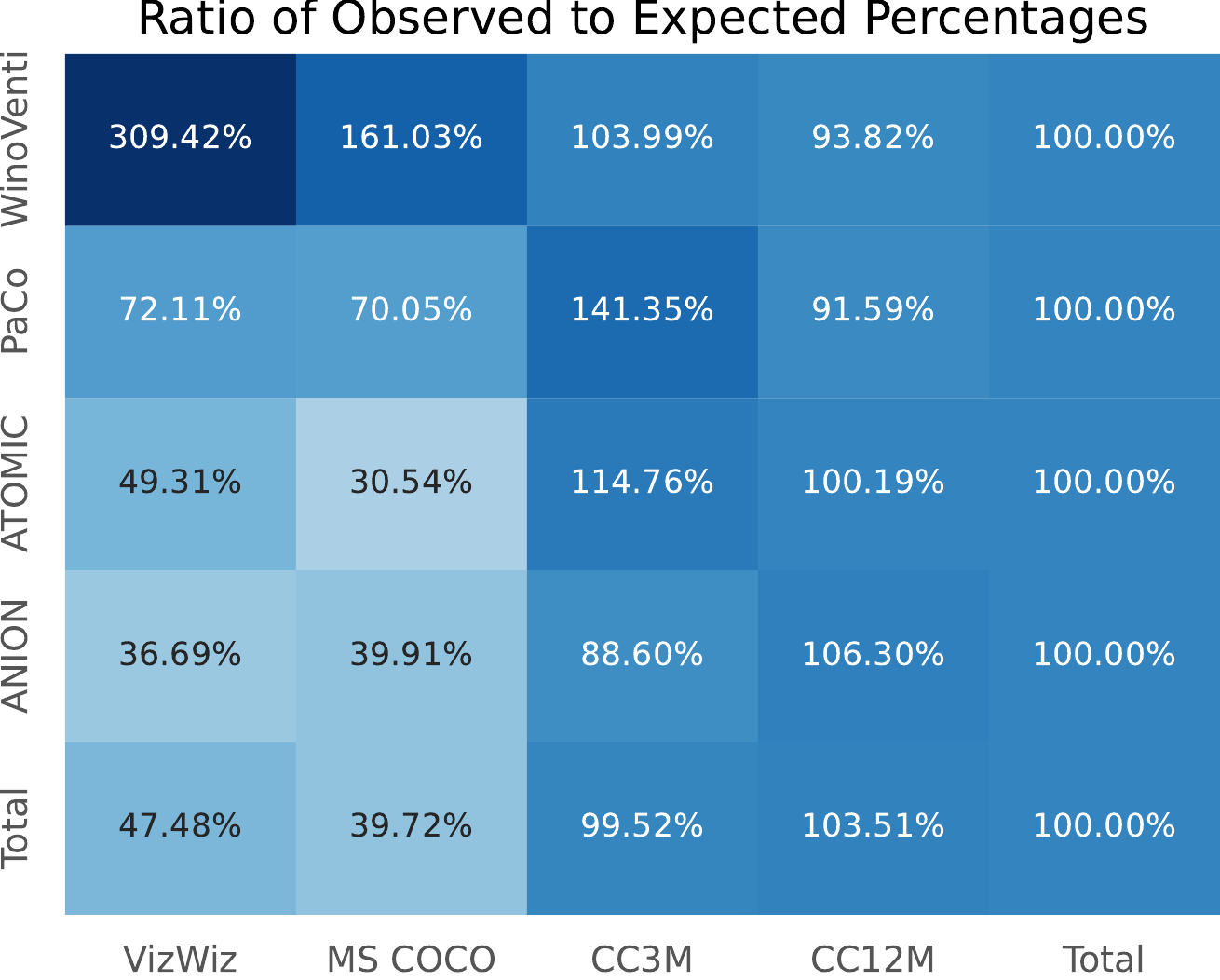}
    \caption{}
    \label{fig:preconditions_ratio}
 \end{subfigure}
 \hfill
 \begin{subfigure}[b]{0.45\textwidth}
    \centering
    \includegraphics[width=.7\linewidth]{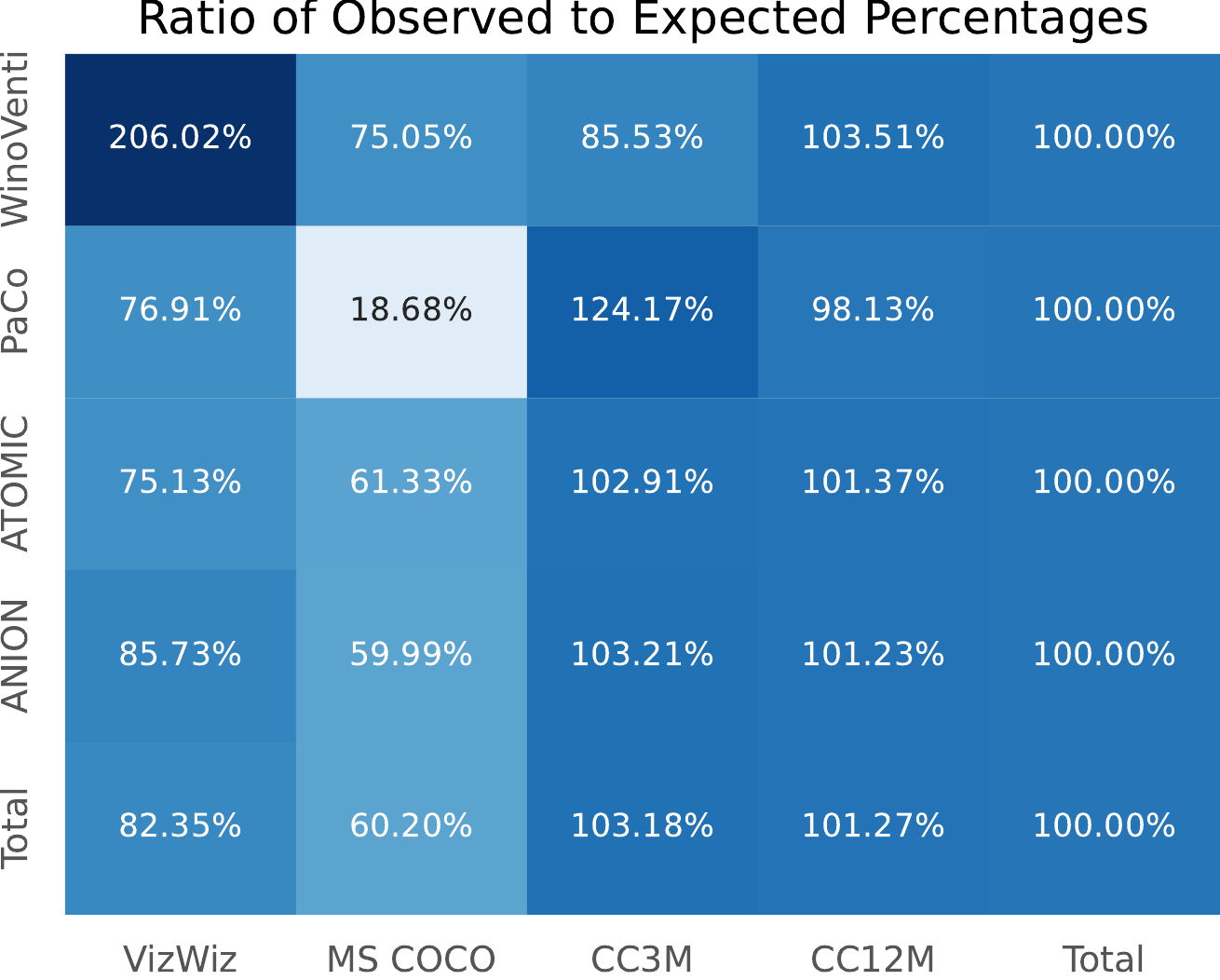}
    \caption{}
    \label{fig:actions_ratio}
 \end{subfigure}
 \hspace*{\fill}
 \caption{\subref{fig:preconditions_percents}) Observed distribution of matches for preconditions. \subref{fig:actions_percents}) Observed distribution of matches for actions. \subref{fig:preconditions_ratio}) Deviation from the expected distribution of matches for preconditions. \subref{fig:actions_ratio}) Deviation from the expected distribution of matches for actions.}
\end{figure*}

\paragraph{Caption Quering Results:} 
With \Cref{fig:preconditions_percents} and \Cref{fig:actions_percents}, we look at where our matches in the CQ method come from. 
This tells us that the majority of our queries come from \textsc{Anion} while the majority of captions come from CC12M, which is unsurprising given they are our largest datasets. \Cref{fig:preconditions_ratio} and \Cref{fig:actions_ratio} take the ratio of the observed percentages to the percentages we would expect based purely on the sizes of the datasets. While we always use every query, the captions can be taken from any of the image datasets. This means that some may be over or underrepresented based on their fit for a given NLI dataset. For example, in \Cref{fig:actions_ratio} it appears that MS COCO captions are not good matches for \textit{PaCo} actions.

We also report more results on \textit{Image Querying} in \Cref{subsec:appendix-image-search}.
\section{Evaluation and Discussion}
\label{sec:vlm-results}

In this section, we focus on the \pvli tasks. 
We first benchmark state-of-the-art visual language models on the inference (\Cref{subsec:pvli-results}) and reasoning tasks (\Cref{subsec:pvlr-for-pvli}). 
Then, we focus on sources of bias in the data through counterfactual analysis in the inference task (\Cref{subsec:bias-results}).

\subsection{Inference Benchmarking Results}
\label{subsec:pvli-results}
Here, as the main results, we benchmark the SoTA VLM models in the PVLI task.

\paragraph{Experimental Setup:} 
We used 4 SOTA vision-language models: ViLBERT~\cite{lu2019vilbert}, ViLT~\cite{kim2021vilt}, FLAVA~\cite{singh2022flava}, and CLIP~\cite{radford2021learning}.


For the ViLBERT~\cite{lu2019vilbert} model, we used the pre-trained model provided by the authors\footnote{https://github.com/facebookresearch/vilbert-multi-task}. For the rest of the models, we use the pre-trained weights from the Hugging Face library~\cite{huggingface}.  
ViLBERT has separate encoders for processing images and text, hence we feed the text and image separately. 
The model then predicts one of the 2 outcomes in the PVLI, that is whether the precondition (image) allows or prevents the commonsense statement.
The ViLBERT model, provided by the authors, is originally fine-tuned on the Visual Natural Language Inference (VSNLI) task~\cite{vu2018grounded}; hence it is familiar with the structure of the PVLI task and we can use it for a zero-shot evaluation as well as the fine-tined evaluation.
We fine-tune ViLBERT on the PVLI training set with a batch size of 32 for 5 epochs, with the Adam Optimiser to optimize the cross entropy loss between the actual and the predicted labels.
For all other hyperparameters, we used the default values by authors.
Finally, we report the accuracy of the ViLBERT model on the PVLI test set in both zero-shot and fine-tuned setups.

The Hugging Face library contains the ViLT~\cite{kim2021vilt} pre-trained on the Visual Question-Answering task, in which the model has to find an answer from a predefined set of tokens including \textit{yes} and \textit{no}. 
So for zero-shot and fine-tuned results, we format the PVLI dataset into a question-answering format with binary \textit{yes/no} answers. 
The statement is converted into a question format by appending the phrase  \QT{Is this possible?} to the statement. This question is then fed into the model along with the associated image, which acts as the premise. 
The model then outputs one of the 2 labels - \textit{yes} or \textit{no}, which we use to compute its accuracy on the task.

FLAVA~\cite{singh2022flava} and CLIP~\cite{radford2021learning} are multi-modal vision and language models that can be used for tasks such as image-text similarity or zero-shot image classification.
Similar to ViLT, the hugging face library does not provide CLIP and FLAVA models, pre-trained on binary or multi-label classification tasks. 
For the fine-tuned results of FLAVA model, we extract the multi-modal embeddings it generates and feed them to a classification head. This classification head is fine-tuned on the VSNLI before using in our experiments.
For the CLIP model, we utilize the similarity scores between the visual and the textual features. 
Similarly, we feed the features through a classification head to output the label which indicates whether the precondition \QT{allows} or \QT{prevents} the common sense statement.
We then report the accuracy of the resulting models on the PVLI tasks in zero-shot and fine-tuned, w.r.t. PVLI, setups.

From the weak supervision data, we randomly sample 16K for \textit{tuning} and 6K as \textit{noisy test} set.
For the \textit{clean test} set we used the $261$ human-verified samples obtained through AMT experiments in \Cref{sec:appendix-amt}.
The experiments are conducted on a commodity workstation with an Intel Xeon Gold 5217 CPU and an NVIDIA RTX 8000 GPU.

\begin{table}[t]
\centering
\small
\resizebox{\columnwidth}{!}{%
\begin{tabular}{l|ll|ll}
  \toprule  
       & \multicolumn{2}{l|}{0-shot} & \multicolumn{2}{l}{Finetuned} \\
  Model& Noisy Test& Clean Test& Noisy Test& Clean Test\\ \midrule
  ViLBERT& 52.02& 48.48 & 78.75 & 55.68 \\
  ViLT& 50.88 & 45.83 & 77.92& 55.68 \\
  CLIP&  30.15 & 42.80 & 73.13 & 56.82 \\
  FLAVA& 47.38 & 53.78 & 80.43 & 59.47 \\
  \bottomrule
  Random& 63.47& 56.08 & &  \\
  \bottomrule
\end{tabular}%
}
\caption{Results of SoTA Visual Language Models on the PVLI task. 
}
\label{tab:pvli-results}
\end{table}

\paragraph{Discussion}
\Cref{tab:pvli-results} summarizes the results of SoTA VLMs on the PVLI task. 
In the zero-shot setup, all the models perform below the random baseline, showing the difficulty and novelty of the task for the models.
After fine-tuning, the models' performance improves above the random guess, where the FLAVA's~\cite{singh2022flava} performance elevates by $33.05$ points of accuracy to $80.43\%$ on the \textit{noisy-test}.
However, it still is not mastering the task.
Overall, this shows that SOTA methods generally fall behind human-level performance, therefore indicating the need for further research in order to improve the comprehension of preconditions by commonsense visual reasoners.

\subsection{Analysis with Fine-tuning}
\label{subsec:curve-results}
In the above evaluation on PVLI, we observe that all models get higher scores after a full fine-tuning process.
Here, we dissect the fine-tuning process to find at what point the model understands the task's requirements.

\paragraph{Experimental Setup}
Here we focus on FLAVA~\cite{singh2022flava} as one of the top-performing models in PVLI. 
We carry the setup from \Cref{subsec:pvli-results} and evaluate FLAVA on the noisy test set in fine-grained intervals during fine-tuning.

\begin{figure}
    \centering
    \includegraphics[width=\columnwidth]{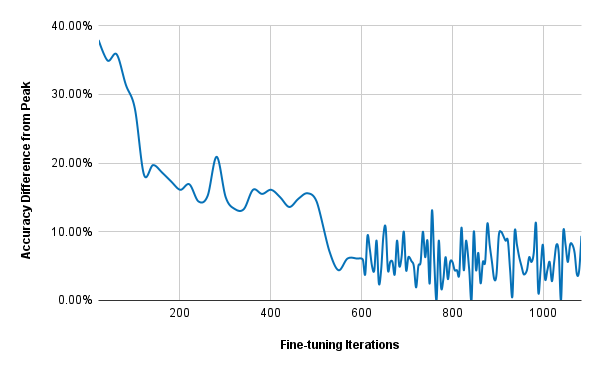}
    \caption{Accuracy difference from the peak value of fine-tuning FLAVA (lower is better) with increasing amounts of tuning data from PVLI. The batch size is 64.}
    \label{fig:tune-curve}
\end{figure}

\paragraph{Discussion}
\Cref{fig:tune-curve} illustrates the progression of the FLAVA model toward its peak accuracy performance. 
As illustrated, the model's performance saturates after 600 iterations of fine-tuning (or 38K instances).
The slow saturation of the accuracy score here suggests that the instances in PVLI are not trivial for the model and it has to see a substantial number of instances to be able to perform the task.
Considering that the FLAVA has been pre-trained on a vast corpus, our result shows the novelty and uniqueness of the PVLI task.
This result is consistent with the similar analysis in \citet{qasemi2022paco}, for comparing MNLI task with PNLI (text-only).

\subsection{Identifying Sources of Bias}
\label{subsec:bias-results}
Large LMs (and by extension VLMs) tend to learn to solve the dataset rather than the task~\cite{le2020adversarial}, by overfitting spurious correlations in the data~\cite{xu2022does}.
To quantify and eliminate such biases in the data/model, recent studies conduct counterfactual inference to debias textual resources used in text classification \citet{qian2021counterfactual} and information extraction tasks \cite{wang2022should,wang2023extracting}. 
Internally, debiasing through counterfactuals works on the model trained or fine-tuned on the biased classification data. 
During inference, this technique creates counterfactuals where parts or all of the input are wiped out to observe what the model would give by seeing only the biasing factors.
In this way, bias terms can be distilled from the model, which can be further deducted from the original prediction for debiasing.
Specifically, \citet{qian2021counterfactual} design two types of counterfactual variations of the input to produce two counterfactual output distributions that model label bias and keyword bias in the model.

\paragraph{Experimental Setup}
Since our data contain both images and text, we modified the counterfactuals in \citet{qian2021counterfactual} to fit the task. 
We create four counterfactual variants of the inputs to consider, visual-token bias, textual-token bias, image bias, and text bias.
In the visual-token bias and textual-token bias, we partially mask the input image ($50\%$) and text ($67\%$ as in \cite{qian-etal-2021-counterfactual}) respectively with no change to the other modality of input.
In the image bias and text bias we blind the model in one modality by fully masking their respective modalities.
Here, we focus on the FLAVA~\cite{singh2022flava} model and carry over the setup from \Cref{subsec:pvli-results} on the noisy test set.

\paragraph{Discussion}
Our results show that the visually blind FLAVA~\cite{singh2022flava} model is performing on par with the original model ($79.88$ accuracy on noisy test). 
This shows that the model may overly rely on the text modality as a shortcut in most of the instances rather than utilizing both image and text.
This result further motivates the need for further research in multi-modal debiasing techniques for both data and models.

\subsection{Utilizing Rationale for Inference Task}
\label{subsec:pvlr-for-pvli}
Here, we try to answer the question \QT{How can the rationales contribute to the inference task?}. 
In other words, we show how the generated rationales can become a piece of useful evidence for inference.

As discussed in \Cref{sec:related-works} (under \QT{Free-Text Rationale Generation}), even though there exists a rich body of literature on the free-text rationale generation models in the text-only tasks, there are limited publicly available models for the visual language tasks. 
We implement the architecture proposed in \citet{ayyubi2020generating} for visually-guided rationale generation\footnote{At the time of this writing, the original implementation of \citet{ayyubi2020generating} was not public}.
The architecture feeds the visual embeddings from a VLM to the decoder of a LM and jointly trains both in an end-to-end fashion.

\paragraph{Experimental Setup}
We do an experiment similar to \Cref{subsec:pvli-results}, except that the VLM model is trained with both the textual \textit{hypothesis} and \textit{rationale} plus the visual \textit{premise} as input.
To contain the length of this experiment we only focus on the FLAVA~\cite{singh2022flava} VLM, and evaluate the performance on the noisy test set in a fully fine-tuned setup.
We separately experiment with two types of rationales as input: the \textit{FLAVA-rationale-gen} gets the generated rationale, and the \textit{FLAVA-rationale-gold} gets the ground-truth rationale from \pvli\@.

For our implementation of \citet{ayyubi2020generating} to generate the rationale, we use a separate FLAVA~\cite{singh2022flava} as the VLM to embed the multi-modal input and use GPT-2~\cite{gpt2} as a decoder-only LM to generated the rationale.
We initialize both models, from pre-trained weights on Hugging Face~\cite{huggingface} library and fine-tune them on \pvli data for the rationale generation task given the input (text and image).

\paragraph{Discussion}
The inference accuracy of the \textit{FLAVA-rationale-gold} and \textit{FLAVA-rationale-gen} is $94.2$ and $80.56$ respectively.
First, the significant jump in the performance of \textit{FLAVA-rationale-gold} (from the base of $80.43$ in \Cref{tab:pvli-results}) shows that in the presence of a competent rationalization model, the generated rationales can significantly contribute to the inference task.
Second, we observe that a rationale model as simple as \textit{FLAVA-rationale-gen}, can also contribute to the performance (although slightly) of the visual preconditioned inference task.
This further motivates the need for further research in multi-modal rationalization models.

\section{Related Works}
\label{sec:related-works}

\paragraph{Preconditions of Commonsense Knowledge}
reasoning with preconditions of \CSn has been studied in the context of affordance in different fields from cognitive sciences~\cite{garbarini2004root} to robotics~\cite{ahn2022can} but was recently brought up in natural language understanding.
In NLP, the focus has been mainly on proposing human-verified learning resources \cite{qasemi2022paco,rudinger2020thinking,hwang2020comet,sap2019atomic,heindorf2020causenet,do2021rotten,jiang2021m}.
Among them, \citet{qasemi2022paco} and \citet{rudinger2020thinking} propose variations of the canonical NLI task for preconditioned inference in common sense. 
\citet{qasemi2022pinks} propose a combination of weak-supervision strategy and biased masking to improve LMs' performance in the task.

\paragraph{Visual Language Inference}
With the advent of visual language models (VLMs; \citealt{li2022uniformer,liu2021cross,li2019visualbert,cho2021unifying,huang2022vlg}) that can simultaneously process visual and linguistic information, there is growing attention to enrich text-only tasks with visual context~\cite{parcalabescu2021valse,xie2018visual,vu2018grounded}.
\citet{vu2018grounded} propose a visually-grounded version of the textual entailment task, supported by the cognitive science view of enriching meaning representations with multiple modalities. 
According to how Visual Language Inference~(VLI; \citealt{xie2018visual,vu2018grounded}) is defined, the task is regarded as a visual extension of the NLI task.
In VLI, the \textit{premise} is substituted with an image with visual context instead of the text in NLI \cite{xie2018visual}.
Instead of relying on crowdsourcing, both works augment the Stanford NLI (SNLI) dataset \cite{bowman2015snli}. 
Since the textual \textit{premise}s of SNLI are extracted from image captions on Flickr, each \textit{premise} can be easily replaced with its respective image.
Our proposed PVLI task is a variation of the proposed VLI that focuses on the preconditions (affordance) of tasks/objects (similar relation exists between general NLI, e.g. MNLI versus the PNLI task).

\paragraph{Weak Supervision}
Instead of using direct supervision from annotated data, weak supervision in NLP tasks typically use linguistic patterns to infer large-scale \QT{noisy} or \QT{imperfect} labels on unlabelled corpora
~\cite{rekatsinas2017holoclean,zhang2017deepdive,dehghani2017neural,singh2022viphy}, e.g.\ using heuristic rules.
Models fine-tuned on weak supervision data have shown considerable improvements across NLU tasks lacking direct supervision, including temporal commonsense reasoning \cite{zhou2020temporal}, rationale generation \cite{brahman2020learning}, document ranking \cite{dehghani2017neural}, and preconditioned inference~\cite{qasemi2022pinks}.
\citet{dai2021ultra, choi2018ultra} use weak supervision from linguistically mined or LM-generated noisy data to enhance ultra-fine entity typing. \citet{choi2018ultra} also uses human annotators to create a small ground-truth test set for testing.

\paragraph{Free-Text Rationale Generation}
There is a large body of research on free-text rationale generation toward faithful and explainable NLP.
Work like this typically fine-tunes a single LM to generate the task output and rationale \cite{narang2020wt5,marasovic2021few,zelikman2022star}, or uses a separate LM to generate the rationale that another LM uses to generate the output \cite{wang2022pinto,wei2022chain,kumar2020nile,rajani2019explain}.
In the visual-language realm, free-text rationale generation is limited, where based on our observation it can be due to the lack of large-scale learning resources. 
\citet{dua2021beyond} and \citet{ayyubi2020generating} repurpose the VCR~\cite{zellers2019vcr} data and propose VL models to generate free-text rationale (instead of picking one as is in the VCR) for it.
Other works, e.g. \citet{su2022language,li2022scene}, use visual inputs for text generation, but they are not focused on the rationale generation. 
\section{Conclusion and Future Work}
\label{sec:conclusion}

 We introduce the Preconditioned Visual Language Inference and Rationalization tasks to measure how SOTA VLMs can extract preconditions and infer the affordance of the objects.
 We propose three strategies for obtaining and retrieving a rich amount of cheap and allowably noisy supervision signals for the task and use them to create a crowd-verified evaluation dataset to benchmark future models.
 Our experiments show that SOTA VLMs largely fall behind human performance in the proposed task raw a road map to address the challenges ahead in improving them.
\section*{Limitations}
\label{sec:limitations}
Image captioning datasets are limited both in breadth and depth.
We have not investigated the use of automatically generated captions, e.g. \citet{wang2022ofa}, in our weak-supervised pipeline, but it is a viable path for future extensions of this work.
Alternatively, automatic text-to-image generation techniques, e.g. stable diffusion~\cite{Rombach_2022_CVPR} or Dall-E~\cite{ramesh2022hierarchical}, are gaining a lot of attention and are promising but require a lot of prompt engineering that is challenging on a large scale.
In addition, the lack of access to a large number of free-text rationale generation models (through libraries such as Huggingface~\cite{huggingface}) limited the evaluation of our PVLR tasks. 
We hope the availability of resources, such as ours, elicits more research effort in the field.

\section*{Ethical Concerns}
We started from publicly available data that is both crowd-verified and neutralized, however, multiple studies have shown the existence of bias and ethical issues in such resources, e.g.~\citet{mehrabi2021lawyers}.
Since our work is based on weak supervision, we have no additional filter on the acquired instances, hence our resource exacerbates the bias in models by reinforcing it with biased evidence, e.g. results from the query \QT{fat person} will only return images of obese white males. 
In addition, there is a combination of well-studied biases in the large models trained on raw text, e.g.~\citet{bender2021dangers}.

Finally, in this work, we have only relied on English resources. In addition, we have only used English-speaking annotators. Hence the judgments and design decisions are heavily skewed culturally which will aggravate the bias issues of our work.


\bibliography{references}
\bibliographystyle{acl_natbib}

\clearpage
\setcounter{page}{1}
	
\appendix
\begin{center}
    {
    \Large\textbf{Appendices}
    }
\end{center}

\section{Weak Supervision Methods}
\label{sec:appendix}

\subsection{Implementation Details and Experimental Setup}
\label{subsec:data-experimental-setup}
This section discusses the experimental setup and implementation details for the results in \Cref{sec:data-analysis}.

\paragraph{Precondition Resources:}
For our P-NLI datasets, we pull from \textsc{Anion} \cite{jiang2021mad}, \textsc{Atomic} \cite{sap2018atomic}, \textit{PaCo} \cite{qasemi2022paco}, $\delta$-NLI~\cite{rudinger2020thinking} and \textsc{WinoVenti} \cite{do2021rotten}.
For the image caption datasets, we use CC12M \cite{changpinyo2021cc12m}, CC3M \cite{sharma2018conceptual}, MS COCO \cite{lin2014mscoco}, and VizWiz \cite{gurari2020vizwiz}.

\paragraph{Preprocessing Setup:}
Since \textsc{Anion} and \textsc{Atomic} use fixed identifiers (Alice/Bob, PersonX/PersonY), we rely on regex rules to replace them with ``the person'' and ``another person''. \textsc{WinoVenti} uses random first names, and so we utilize Flair's \textit{ner-english-fast} model \cite{akbik2018coling} to identify and replace the spans that are identified as people with greater than 90\% confidence. \textit{PaCo} does not have any such identifiers to replace.

For the image caption data, we break the captions into multiple lines using Natural Language Toolkit's sentence tokenizer \cite{bird2009nltk} in combination with splitting on newline characters.
We also notice that some contained ``<PERSON>'' tags and use regex to replace them.

As the last step, we leverage regex to fix whitespace issues and replace ``the person's'' with ``their'' to increase fluidity. 
We also found that datasets were easier to clean after lowercasing, particularly as some contained inconsistent capitalization.

\paragraph{Extraction from Captions Setup:}
We modify some of the original labeling functions from \textit{PInKS}~\cite{qasemi2022pinks} and add eight new ones after inspecting our caption corpus. In \textit{PInKS}, the authors also calculate precision values for each of their labeling functions by sampling 20 examples from each function.
The samples are then marked as \textit{relevant} (score of 1) or \textit{irrelevant} (score of 0) to the task by two human annotators.
The average score of each labeling function provides an estimate of the quality that each labeling function returns and is used for tie-breaking matches or filtering out low-quality functions.
We follow their lead and do the same, computing these precision values specifically on our caption corpus.
\Cref{tab:lf-patterns} summarizes all the labeling functions, patterns, their precision, and other details associated with them.
\begin{table*}[t!]
    \centering
    \small
    \begin{tabular}{llll}
    \toprule
    \textbf{\normalsize Label} & \textbf{\normalsize Conjunction} & \textbf{\normalsize Precision} & \textbf{\normalsize Regex Pattern} \\
    \midrule
    enables & so that & 0.689 & \{P\} so that \{A\} \\
            & in order to & 0.650 & \{P\} in order to \{A\} \\
            & because & 0.625 & \{A\} because (?!of$\backslash$b)\{P\} \\
            & \textbf{due to} & 0.550 & \{A\} due to \{P\} \\
            & in case & 0.475 & \{A\} in case (?!of$\backslash$b)\{P\} \\
            & as if & 0.400 & \{A\} as if \{P\} \\
            & as long as & 0.375 & \{A\} as long as \{P\} \\
            & if & 0.150 & \{A\}(?<!$\backslash$bas) if  (?!not$\backslash$b)\{P\}\\
            & in the event & 0.100 & \{A\} in the event \{P\}\\
            & on condition & 0.045 & \{A\} on condition (?!of anonymity$\backslash$b)\{P\}\\
            & supposing & 0.000$^\ast$ & \{A\} supposing \{P\} \\
            & on the assumption & 0.000$^\ast$ & \{A\} on the assumption \{P\}\\
            & in the case that & 0.000$^\ast$ & \{A\} in the case that \{P\}\\
            & contingent upon & 0.000$^\ast$ & \{A\} contingent upon \{P\}\\
            & with the proviso & --- & \{A\} with the proviso \{P\}\\
            & to understand event & --- & to understand the event "\{E\}", it is important to know that \{P\}$\backslash$. \\
            & statement is true & --- & the statement "\{E\}" is true because \{P\}$\backslash$.\\
            & only if & --- & \{A\} only if \{P\}\\
            & on these terms & --- & \{A\} on these terms \{P\}\\
            & makes possible & --- & \{P\} makes \{A\} possible$\backslash$.\\
    \midrule
    disables & unless & 0.750 & \{A\} unless \{P\} \\
             & even though & 0.550 & \{A\} even though \{P\} \\
             & despite & 0.475 & \{A\} despite \{P\} \\
             & if not & 0.300 & \{A\}(?<!$\backslash$bas) if not (?!(more|most|many|all)$\backslash$b)\{P\} \\
             & without & 0.257 & \{A\} without \{P\}\\
             & but & 0.175 & \{A\} but \{NP\}\\
             & except & 0.075 & \{A\} except \{P\}\\
             & lest & 0.045$^\ast$ & \{A\} lest \{P\}\\
             & excepting that & --- & \{A\} excepting that \{P\}\\
             & except for & --- & \{A\} except for \{P\}\\
    \bottomrule
    \end{tabular}
    \caption{Regex patterns for the labeling functions. A=action, E=event, P=precondition, NP=negative precondition. Patterns with fewer than 20 examples in the corpora are marked with asterisks, and those with no examples are left empty. Bolded conjunctions were followed with part-of-speech tagging to confirm that they were used as conjunctions. 
    }
    \label{tab:lf-patterns}
\end{table*}
Balancing quality and quantity in our data, we select a threshold of $0.60$ and only use the labeling functions that meet this minimum.
The labeling functions are applied using Snorkel \cite{ratner2017snorkel}, a SOTA framework for algorithmically labeling data---see the original \textit{PInKS} paper for more detail on the setup for Snorkel.
Finally, We noticed that not all of our patterns were being used in the sentence as conjunctions, and utilized Flair's \textit{pos-english-fast} model to remove some examples for select patterns.

\paragraph{Caption Querying Setup:}
For our models, we use the sentence transformers \cite{reimers-gurevych-2019-sentence} all-distilroberta-v1, all-MiniLM-L12-v2, and all-mpnet-base-v2 from HuggingFace~\cite{huggingface}.
When forming the rankings, we retrieve the 50 closest captions, as that provides a decent overlap and completes within a reasonable amount of time. To aggregate our rankings and select the best caption, we use Copeland's method~\cite{copeland1951reasonable}. 
To compare our rankings for model agreement, we utilize the extrapolated form of rank-biased overlap~\cite{webber2010rbo}. 

\begin{figure}[ht]
    \centering
    \includegraphics[width=\linewidth]{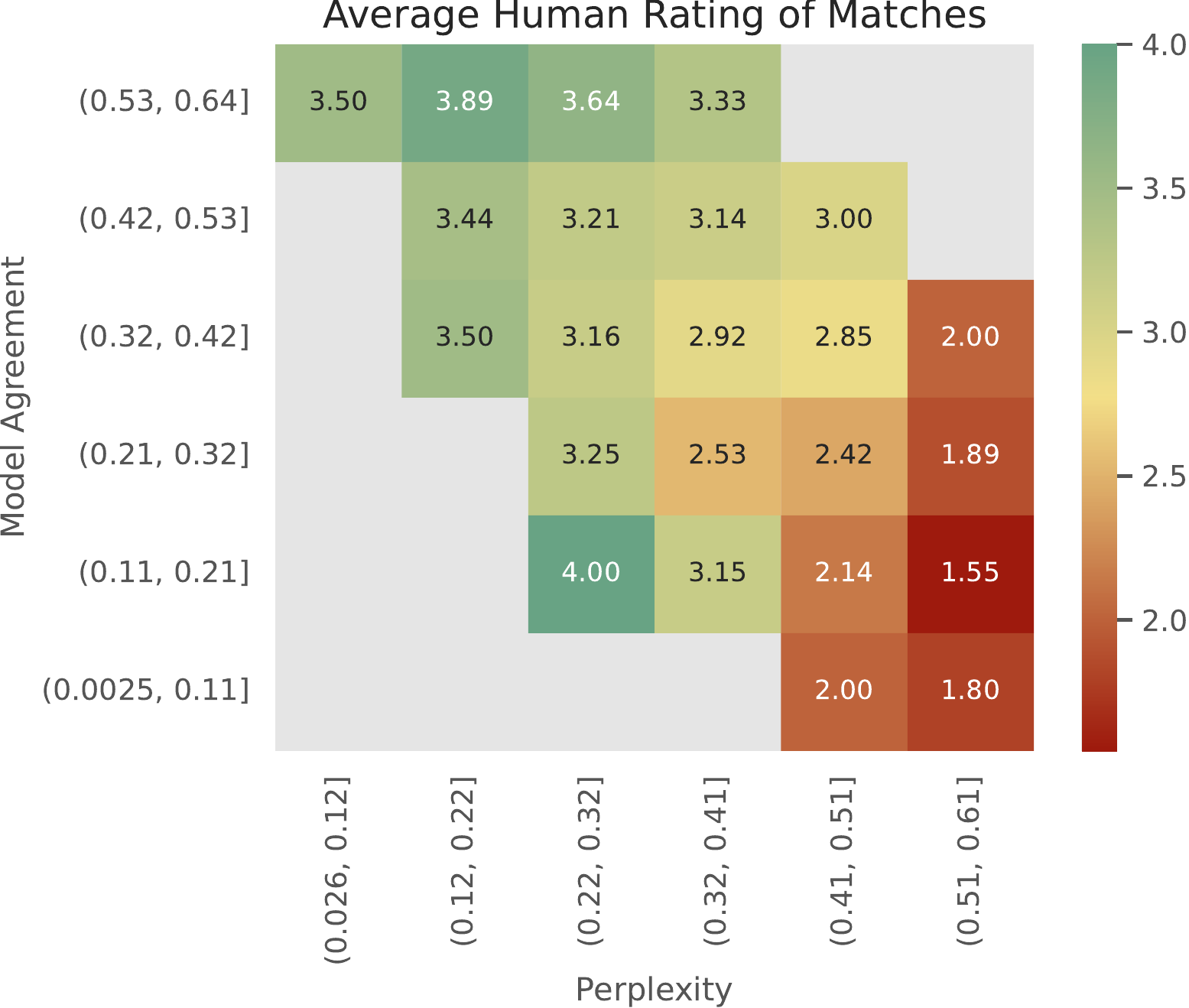}
    \caption{Heatmap graph comparing the measures of perplexity and model agreement with expert human evaluation in the caption querying method. Bins are computed using 6-quantiles for each axis.}
    \label{fig:perplexity_agreement_heatmap}
\end{figure}

\begin{figure}[ht]
    \centering
    \includegraphics[width=\linewidth]{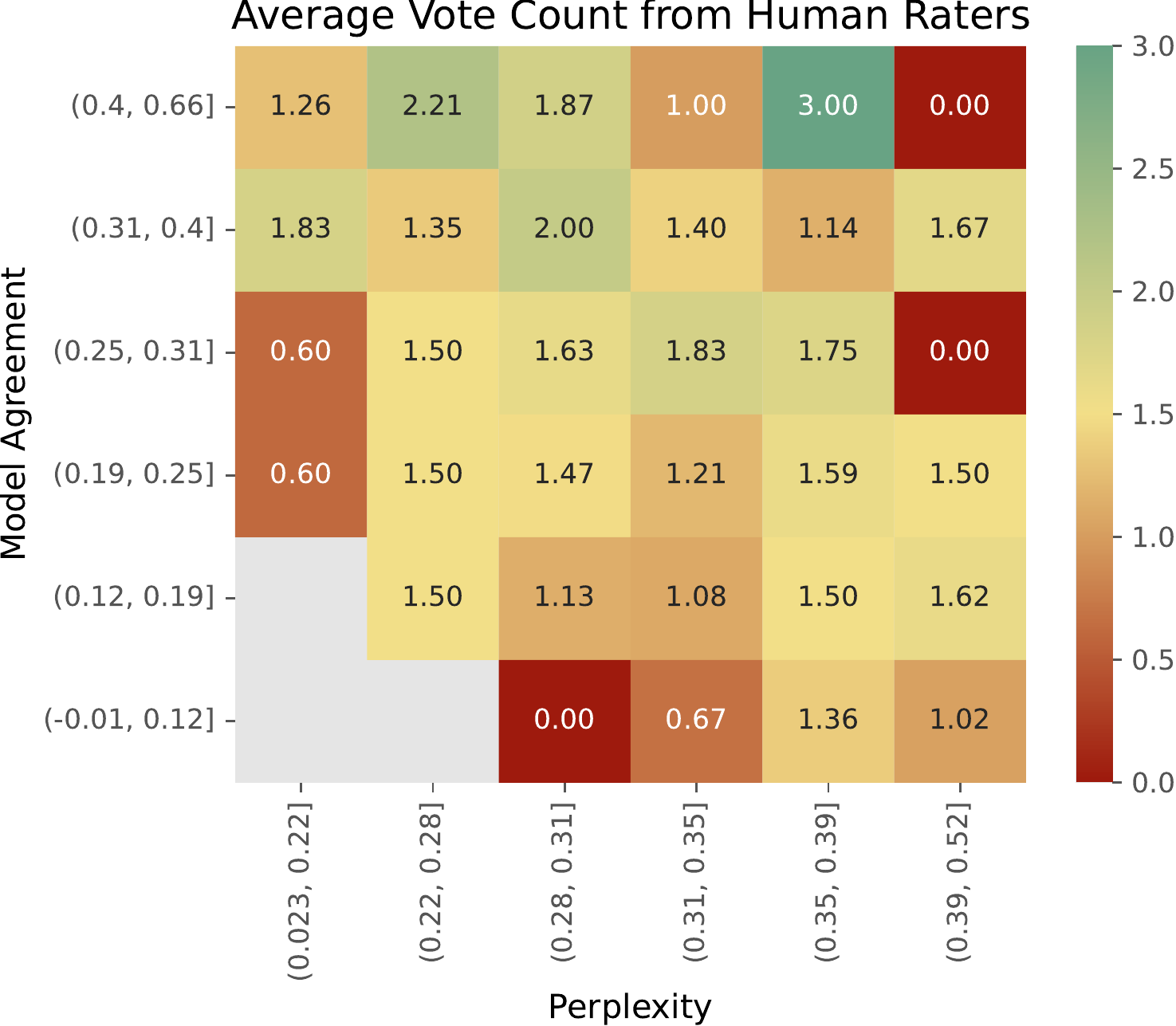}
    \caption{Heatmap graph comparing the measures of perplexity and model agreement with human evaluation from Amazon Mechanical Turk in the caption querying method. Bins are computed using 6-quantiles for each axis. 
    }
    \label{fig:perplexity_agreement_heatmap_mturk}
\end{figure}

Furthermore, we annotate a subset of our data to assess the ability of perplexity and model agreement to separate good training examples from poor ones. 
When we rate examples for quality of match between the statement (precondition/action) and the fetched image caption, on a scale from 1 (worst) to 4 (best), we find that our measures are reasonably successful (See~\Cref{fig:perplexity_agreement_heatmap}). 
However, when we ask Amazon Mechanical Turk workers to vote on examples for overall quality, requiring that both statements and images be cohesive, the measures are unable to isolate better examples (See~\Cref{fig:perplexity_agreement_heatmap_mturk}). 
Given that the measures are based purely on the textual match, it makes sense that it would perform better without the incorporation of the image. 
Unfortunately, matching with a caption is not always sufficient for matching with the associated image. 
Further work is needed to develop useful heuristics for the overall quality of a training example.

\paragraph{Image Querying Setup:}
To find the top images on the internet, we use \textit{Google Images Download} \footnote{\url{https://github.com/Joeclinton1/google-images-download}} to retrieve the URLs of images.
We obtain the top 10 images for each query as it is large enough to generate lots of data while keeping them relevant to the query.


\subsection{Image Search Results}
\label{subsec:appendix-image-search}
While less information is available for the \textit{Image Querying} data, we can look at the websites most frequently drawn from for the matches. \Cref{tab:preconditions-websites} and \Cref{tab:actions-websites} display the top 10 websites for each NLI dataset for preconditions and actions, respectively. If desired, it is possible to remove images from unwanted websites.

\begin{table*}[t!]
\centering
\begin{tabular}{lll}
\toprule
                      \textsc{WinoVenti} &                       \textit{PaCo} &                        \textsc{Anion} \\
\midrule
        m.media-amazon.com (89) &       quotefancy.com (102) &        quotefancy.com (4721) \\
               i.ytimg.com (85) &           i.ytimg.com (80) & thumbs.dreamstime.com (2668) \\
           cdn.shopify.com (67) & thumbs.dreamstime.com (73) &             i0.wp.com (2074) \\
      upload.wikimedia.org (64) &             i0.wp.com (72) &          i.pinimg.com (1662) \\
     media.istockphoto.com (52) & media.istockphoto.com (46) &       www.wikihow.com (1597) \\
     thumbs.dreamstime.com (50) &    m.media-amazon.com (46) &  www.verywellmind.com (1546) \\
                 i0.wp.com (47) &          i.pinimg.com (39) & media.istockphoto.com (1251) \\
images.squarespace-cdn.com (35) &          c8.alamy.com (36) &        miro.medium.com (997) \\
      i5.walmartimages.com (34) &  upload.wikimedia.org (35) &      www.incimages.com (967) \\
              c8.alamy.com (33) &       www.wikihow.com (33) &     previews.123rf.com (900) \\
\bottomrule
\end{tabular}
\caption{Top 10 websites for preconditions by NLI dataset. There are a total of 10,975 unique websites for 50,729 unique images belonging to 82,740 examples.}
\label{tab:preconditions-websites}
\end{table*}

\begin{table*}[t!]
\centering
\begin{tabular}{lll}
\toprule
                 \textsc{WinoVenti} &                            \textit{PaCo} &                        \textsc{Anion} \\
\midrule
   m.media-amazon.com (54) &                  i0.wp.com (79) & thumbs.dreamstime.com (3164) \\
          i.ytimg.com (25) &       www.verywellmind.com (62) &        quotefancy.com (2943) \\
 i5.walmartimages.com (24) &       upload.wikimedia.org (45) &             i0.wp.com (2050) \\
thumbs.dreamstime.com (21) &        post.healthline.com (34) &          c8.alamy.com (1964) \\
      cdn.shopify.com (21) &         media.cheggcdn.com (24) & media.istockphoto.com (1962) \\
            i0.wp.com (20) &             quotefancy.com (19) &       www.wikihow.com (1469) \\
         c8.alamy.com (20) &       qph.cf2.quoracdn.net (19) &  www.verywellmind.com (1262) \\
     i.etsystatic.com (19) &          www.helpguide.org (17) &         i.insider.com (1213) \\
 upload.wikimedia.org (18) &             media.self.com (15) &    previews.123rf.com (1075) \\
media.istockphoto.com (17) & images.squarespace-cdn.com (15) &           i.ytimg.com (1050) \\
\bottomrule
\end{tabular}
\caption{Top 10 websites for actions by NLI dataset. There are a total of 9,700 unique websites for 48,305 unique images belonging to 80,170 examples.}
\label{tab:actions-websites}
\end{table*}

\subsection{Model Sizes and Run-times}
For results in \Cref{tab:pvli-results}, the runtimes are FLAVA=4hr, VilBERT=4hr, Clip=5hr, ViLT=4hr; the model sizes for VLMs and LMs are identical to their respective implementations from the source (e.g. gpt2 has 1.5B parameters on \citet{huggingface}).
The classification head added to VLMs (e.g. FLAVA) has $1.4k$ parameters.

\section{Data Annotation Details}
\label{sec:appendix-amt}

We used Amazon Mechanical Turk (AMT)~\cite{crowston2012amazon} to evaluate the quality of extracted PVLIR instances through our proposed weak supervision methods.
This enabled us to coordinate the study and access a large pool of English-speaking participants as our study population.
The AMT is especially suitable for this study as it can facilitate accessing a diverse population of participants which is necessary for any notion of common sense.
Our study on AMT consists of two parts: a tutorial, which also serves as a qualification test, and the main survey.
In addition, we implemented two levels of quality control: in the first one we use a response checker code and in the second we use human annotators to ensure only high-quality responses wind up in the final data.

\subsection{Main AMT Survey}
\label{subsec:main-amt-survey}

\begin{figure}
    \centering
    \includegraphics[width=0.8\linewidth]{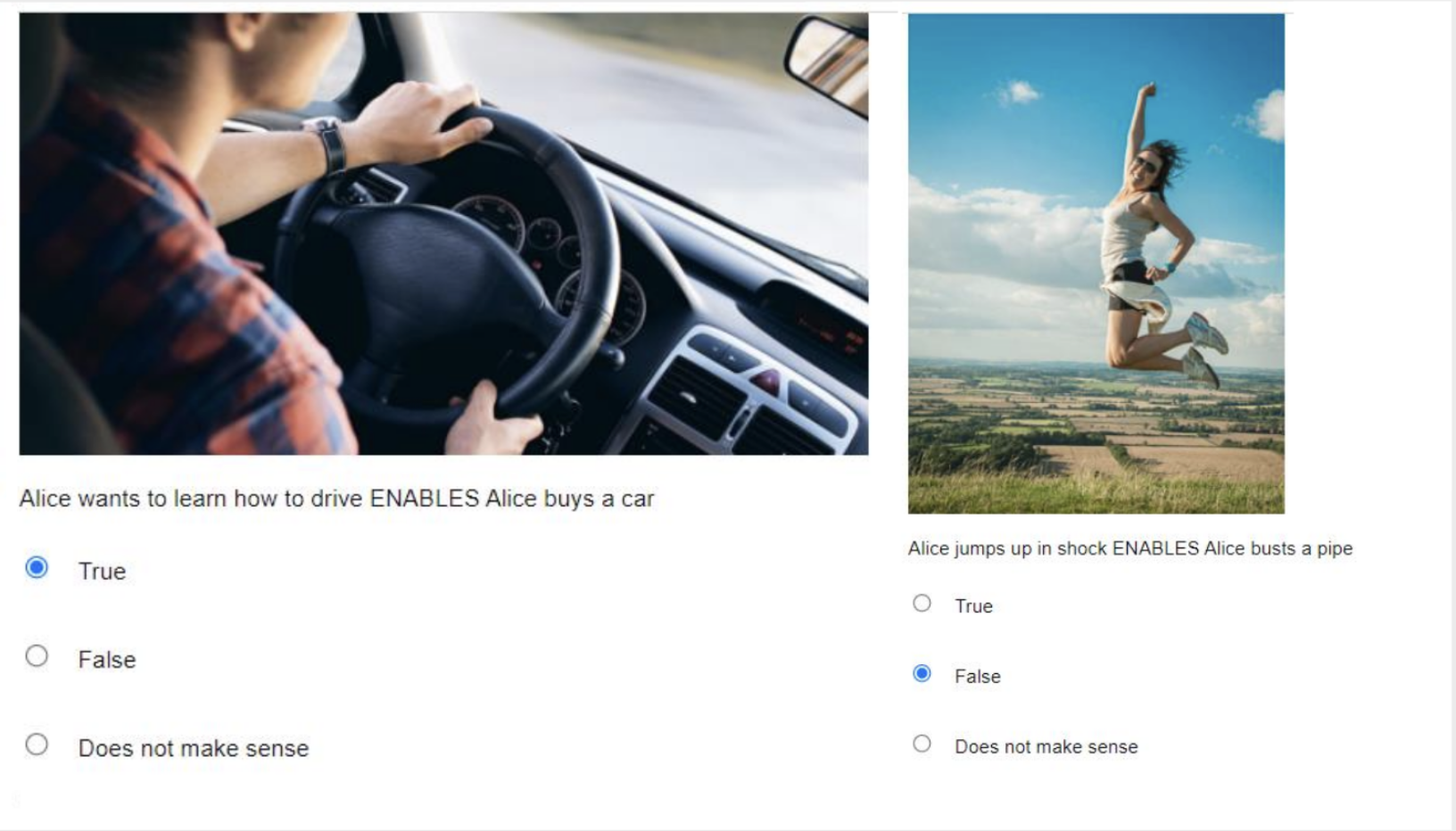}
    \caption{A sample question-unit used in the main survey on the AMT.}
    \label{fig:question-unit}
\end{figure}

In the main survey, the participants are given a set of question units each consisting of a prompt question, an image, and the radio buttons with three options. We then ask participants to select their responses for each prompt question from the available options in the unit (e.g. “true” “false” “not sure” sample in Fig.~\ref{fig:question-unit}). 
We create a question until through the PVLI instances with image and text, that was discussed in \Cref{sec:dataset}.

Since our annotated images are not perfect, there are a lot of possible points of failure that can render the question units to be impossible to understand. For example, some of the annotations may not be correct, the automatic conversion of meta-data to a sentence can be wrong in corner cases, or the image links be corrupted. Hence some of the question units may have odd grammar (e.g. "An net is used for catch fish"). Consequently, some of the question units may be hard to understand or just be wrong. To help us find those question units and ignore them in future iterations, each question unit has a checkbox in front of it with the label "not sure/does not make sense". The participant may choose to select the option and skip answering that prompt. To make the payment structure fair for the participants, they will get paid regardless of their responses. We keep the right reserved to block the participants who abuse this option using the annotator agreement metric.

\subsection{Qualifying Participants}
\label{sec:appendix:quality}
In the tutorial, first, we have prepared detailed instructions that explain to the participants what they need to do and what are the criteria for a good vs bad response. For example, in the instructions, we ask participants to avoid answering “correct”/“incorrect” when they are not sure or when there is something wrong with the image or text of the question unit. The instruction is <500 words with an expected reading time of <7 mins. Additionally, we have prepared a set of good/bad examples associated with each rule that can also be accessed in the tutorial. Each one of the good/bad examples comes with a short explanation that discusses the reason for the good/bad rating of the response. The participants are then asked to give the qualification test as a check on whether they have read and understood the instructions. The qualification test contains $\sim$10 question units similar to the ones they will see in the original survey (due to AMT limitations the qualification question units have a different visual layout but contain the same information). We have carefully designed each qualification question unit such that it tests the participants' understanding of the rules individually and give them feedback on their wrong answers. For example, for the rule discouraging the use of “correct”/“incorrect” when the question unit is invalid, we have two question units where first the image is not visible, and second, the text is gibberish. After successfully passing the test, participants with acceptable scores are granted a qualification badge that allows them to engage in the main survey. It must be noted that the detailed instructions and the good/bad examples are both available in the main survey as a memory refresher for the participants.

To coordinate the study and access a large pool of participants we use Amazon's Mechanical Turk (AMT) service to hire English-speaking people with no specific background as our study population. As part of AMT's service design,  the main survey can be divided into thousands of micro-tasks that each is related to a handful of unique question units. In this setup, the participants may choose their amount of participation in the study by accepting micro-task jobs whenever they want or fits with their schedule. Our goal is that each micro-task takes a short time to complete (less than 1 min) so we can attract a larger group of participants. It must be noted that participants can quit at any time and they will be compensated for their submitted work up until that point. To ensure the quality of the responses, the AMT service allows us to review and accept the responses from each participant individually, this allows us to pinpoint workers with low-quality responses (e.g. disagreement on more than 50 percent of the tasks with other participants) and ban them from future participation. Even after being banned, the participants with low-quality responses will be compensated for their previous accepted works.

\subsection{Mechanical Turk Results}
Asked annotators to go through 500 instances of PVLI. 
Each instance was annotated by 3 randomly-selected workers from mainly English-speaking countries: U.S., Canada, England, India, and Australia. 
We selected the instances that are found correct by at least 2 annotators and use them as the \textit{clean-test} set.
The final \textit{clean-test} set, consists of 261 human-verified instances with 151 allow labels.
The inter-annotator agreement (Fleiss’ Kappa) measure between our annotators is 0.78, showing good agreement among them.






\end{document}